\documentclass[runningheads]{llncs}
\usepackage{graphicx}
\usepackage{stackengine} 
\stackMath
\usepackage{float}
\usepackage{graphicx}
\usepackage{subcaption}
\usepackage{xspace}
\usepackage{amsmath}
\usepackage{makecell}
\usepackage{svg}
\usepackage{amsmath}
\usepackage{tikz}
\usepackage{paralist}
\usepackage{algorithmic}  
\usepackage{algorithm}
\usepackage[algo2e]{algorithm2e}
\usepackage{xkcdcolors}
\usetikzlibrary{shapes,arrows,chains,shadows,positioning}
\usetikzlibrary{arrows.meta}
\usetikzlibrary{decorations.pathreplacing}
\usepackage{geometry}
\usepackage{textcomp}
\usepackage{pgfplots}
\pgfplotsset{width=10cm,compat=1.9}

\usepackage[utf8]{inputenc}
\usepackage[T1]{fontenc}    
\usepackage{hyperref}       
\usepackage{url}            
\usepackage{multirow}
\DeclareUnicodeCharacter{2212}{-}
\begin{document}
\title{ChemTab: A Physics Guided Chemistry Modeling Framework}
\titlerunning{ChemTab}
%
\author{Amol Salunkhe\inst{1}\and
Dwyer Deighan\inst{1}\and
Paul E. DesJardin\inst{1}\and
Varun Chandola\inst{1}
}
\authorrunning{Salunkhe et al.}
%
\institute{University at Buffalo, Buffalo NY 14260, USA 
\email{\{aas22,dwyerdei,ped3,chandola\}@buffalo.edu}}
\maketitle              
\begin{abstract}
Modeling of turbulent combustion system requires modeling the underlying chemistry and the turbulent flow. Solving both systems simultaneously is computationally prohibitive. Instead, given the difference in scales at which the two sub-systems evolve, the two sub-systems are typically (re)solved separately. Popular approaches such as the {\em Flamelet Generated Manifolds} (FGM) use a two-step strategy where the governing reaction kinetics are pre-computed and mapped to a low-dimensional manifold, characterized by a few reaction progress variables (model reduction) and the manifold is then ``looked-up'' during the run-time to estimate the high-dimensional system state by the flow system. While existing works have focused on these two steps independently, we show that joint learning of the progress variables and the look-up model, can yield more accurate results. We propose a deep neural network architecture, called {\em ChemTab}, customized for the joint learning task and experimentally demonstrate its superiority over existing state-of-the-art methods.
%
%

\end{abstract}
\section{Introduction}
Modeling of turbulent flow combustion is central in the development of new combustion technologies in aviation, automotive and power generation~\cite{Giusti:2019}. Turbulent flow combustion combines two nonlinear and multi-scale phenomena: {\em turbulent flow} and {\em chemical reactions}. This coupling of the kinetic chemical reaction equations with the set of Navier–Stokes flow equations results in a problem that is too complex to be solved, at full resolution, by the current computational means. Even for a simple fuel such as methane, the combustion chemistry mechanism involves 53 species and 325 chemical reactions~\cite{grimech}, and the numbers increase with increasing fuel complexity. Solving the details of such mechanisms during the flow simulation can consume up to 75\% of the solution time ~\cite{ElAsrag2013ACB}.  

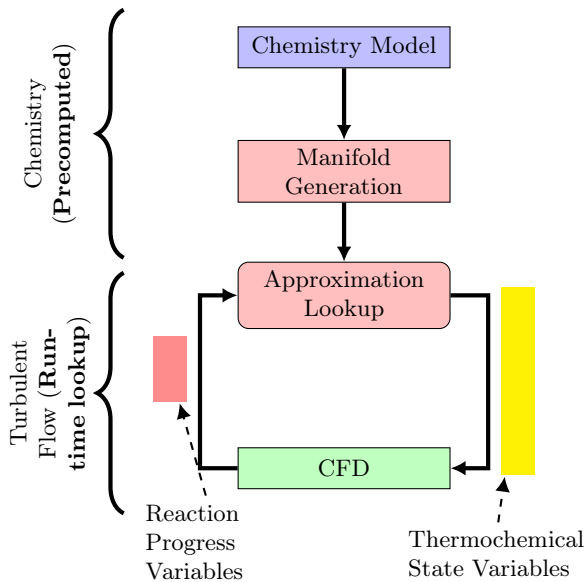
\begin{figure}[h]
  \centering
  \begin{tikzpicture}
  \tikzset{
    >={Latex[width=2mm,length=2mm]},
    base/.style={draw, on grid, align=center, minimum height=4ex},
    proc/.style={base, rectangle, text width=8em},
    term/.style={proc, rounded corners},
  }
  \node[proc,fill=blue!25] (fgm) {Chemistry Model};
  \node[proc,fill=red!25,below of=fgm,yshift=-2em] (mg) {Manifold Generation};
  \node[term,fill=red!25,below of=mg,yshift=-2em] (lib) {Approximation Lookup};
  \node[proc,fill=green!25,below of=lib,yshift=-4em] (cfd) {CFD};
  \draw[->,ultra thick]             (fgm) -- (mg);
  \draw[->,ultra thick]             (mg) -- (lib);
  \draw[->,ultra thick] (cfd.west) -- ++(-0.5,0) -- ++(0,2.3) -- 
  node(small)[xshift=-2em,text height=0.7em,yshift=-3em,fill=red!45,rotate=90,text width=2em]{} 
  (lib.west);
  \draw[<-,ultra thick] (cfd.east) -- ++(0.5,0) -- ++(0,2.3) -- 
  node(full)[text width=7em,text height=0.7em,fill=yellow,rotate=90,yshift=-2em,xshift=-3.5em,align=center] {} (lib.east);
  \draw [decorate,decoration={brace,amplitude=10pt},ultra thick,xshift=2pt]
  (-3,-2.8) -- (-3,0.5) node [proc,rotate=90,draw=none,black,midway,yshift=3em] 
  {Chemistry ({\bf Precomputed})};
  \draw [decorate,decoration={brace,amplitude=10pt},ultra thick,xshift=2pt]
  (-3,-6.2) -- (-3,-3) node [proc,rotate=90,draw=none,black,midway,yshift=3em] 
  {Turbulent Flow ({\bf Run-time lookup})};
  \node[text width=7em,below of=full,xshift=-1em,yshift=-4em](fulltext){Thermochemical State Variables};
  \node[text width=6em,below of=small,xshift=2em,yshift=-4em](smalltext){Reaction Progress Variables};
  \draw[->,dashed,thick] (fulltext) -- (full);
  \draw[->,dashed,thick] (smalltext) -- (small);
\end{tikzpicture}
  \caption{(Re)solving systems separately}
  \label{fig:reduced}
\end{figure}

In most cases, the large scale separation between the combustion chemistry/flame (typically sub millimeter /microsecond scale) and the characteristic turbulent flow (typically centimeter or meter/minute or hour scale) allows simplifying assumptions to be made that enable increased computational efficiency by (re)solving chemistry and flow separately ~\cite{peters2001}.  
In this paper, we focus on approximate methods that deal with handling the chemistry, and in particular, the methods based on laminar flames~\cite{PETERS1984319}. Here the 1-D or single-species flame reactions are solved {\em a priori} and stored. During the flow simulation, these reactions are looked-up to estimate the high-dimensional thermochemical state of the system, as shown in Figure~\ref{fig:reduced}.

Most models developed for increased computational efficiency rely on the existence of a theoretical low-dimensional thermochemical state-space manifold to which the combustion chemistry can be mapped~\cite{maas1992}. The central question then is, {\em how to efficiently model low-dimensional thermochemical manifolds that capture the relevant physics of the problem; and parametrize and approximate these manifolds which can then be accessed during turbulent flow simulations?} 

While existing approaches (collectively referred to as {\em state-space parametrization}~\cite{peters2001,piercemoin2004}) have been successful, they have primarily solved the two sub-problems - {\em progress variable generation} to characterize the manifold, and {\em manifold approximation} to perform the lookup during run-time, independently. This can result in sub-optimal solutions because the progress variables, learnt using methods such as {\em Principal Component Analysis} (PCA)~\cite{sutherland20091563,biglari20154025}, are not necessarily optimized to perform the run-time lookup.  Similarly, while the traditional lookup approaches that use tabulation, or the recently proposed neural network based data-driven alternatives~\cite{bhalla2019}, facilitate efficient look-ups, the construction of the underlying data-structure or machine learning based model is not informed by the learning of the progress variables.

Our main hypothesis is that by simultaneously learning the progress variables and the manifold approximation (lookup model), we can achieve higher accuracy in terms of the estimation of the thermochemical state at run-time. But how does one combine the progress variable learning, an inherently linear mapping task, with a highly non-linear lookup model, while ensuring that the components influence each other during the learning phase? To that end, we propose a framework called {\em ChemTab}, in which the learning of these two components is formulated as a joint optimization task. An implementation of {\em ChemTab}, using a novel deep learning architecture, is proposed. The joint optimization includes a set of mathematical constraints that ensure that the progress variable learning is approximately similar to a PCA-type linear reduction, and, at the same time, can also predict the thermochemical state using a non-linear predictive component. 
The deep learning implementation of ChemTab is shown to reduce the error by 73\%, when compared to an existing tabulation based framework, in predicting one of the key thermochemical term, {\em source energy}, when applied to flames data for a Methane-Air fuel-oxidizer combination generated using the GRI-Mech 3.0 simulator. Moreover, the proposed architecture of ChemTab is shown to outperform a recently proposed state-of-art decoupled PCA+neural network based solution by 24\%.
%
%
\section{Related Work}\label{sec:relatedwork}
In this section we provide a brief overview of existing in low-dimensional thermochemical manifold modeling, focusing more on data-driven methods. We note that there have been works that use physics-driven machine learning models for solving other physics problems ~\cite{willard2020,karpatne2017}, however, these methods generally focus on simpler physics and are not necessarily applicable in the domain of turbulent combustion.

Common approaches to low-dimensional thermochemical manifold modeling are combustion chemistry mechanism reduction and thermochemical state-space parametrization ~\cite{Rastigejev13875} ~\cite{sutherland20091563}. Chemistry mechanism reduction approach cannot be generalized and in the recent past state-space parametrization approach has been the most dominant method comprising of two phases {\em progress variable generation} and {\em manifold approximation}. For progress variable generation, existing methods have either used domain models or numerical methods. 

Domain models like steady Laminar Flamelet Method (SLFM) ~\cite{PETERS1984319}, Flamelet-Generated Manifold (FGM) ~\cite{fgm2000} ~\cite{van2001}, Flamelet/Progress Variable approach (FPVA) ~\cite{piercemoin2004} ~\cite{ihme2005}  and Flamelet-Prolongation of ILDM model (FPI) ~\cite{gicquel2004} theorize that a multi-dimensional flame can be considered as an ensemble of multiple one-dimensional locally laminar flames (flamelets). These flamelets are patametrized by a combination of conserved and reactive scalars ~\cite{fgm2000} ~\cite{van2001} ~\cite{piercemoin2004} ~\cite{bojko2016}. A lot of research in this area builds on the principles laid out in ~\cite{ihme20127715} for progress variables regularization however the fundamental problem of generating adequate number of progress variables that capture the underlying physics is still open.

Numerical methods, like PCA, have shown significant promise for parametrization of the thermochemical state. PCA provides a method of generating reaction progress variables using the flamelet solutions, the state-space variables are still nonlinear functions of the reaction progress variables, and a nonlinear regression is learned to approximate the state-space manifold ~\cite{sutherland20091563} ~\cite{biglari20154025}~\cite{sutherland20091563} ~\cite{malik201830}~\cite{malik2020}. This purely numerical parametrization lack interpretability and may also not be generalizable enough due to variation capture maximization that may overlearn the numerical errors in the data. Linear Autoencoders have also been suggested ~\cite{Yellapantula2021} however this definition lacks incorporation of a principled approach to progress variable generation and thus may not be generalizable.

While domain based model have traditionally relied on tabular lookup, these are not scalable. tabulated data occupies a larger portion of the available memory on every node where the flow simulation is computing. Also the searching and retrieval of this pre-tabulated data becomes increasingly expensive in a higher-dimensional space. For example, assuming a
standard 3 progress variable discretization (200, 100, 50) with say 15 tabulated thermochemical state variables, we obtain a pre-computed combustion table of 120Mb. The addition of a variable such as {\em enthalpy} with a very coarse discretization of 20 points, brings the size of the table to 2.4 Gb. To address the tabulation problem researchers like~\cite{bhalla2019} ~\cite{zhang20201} build on the work of \cite{ihme2009} to investigate the use of a neural networks for manifold approximation which replaces the Tabulation. The mapping between the progress variables (reduced dimensionality) and thermochemical state variables obtained using the flamelets solutions is learnt using a neural network. However, due to the highly non-linear, knotted and discontinuous nature of the lower dimensional manifolds formed by the progress variables generated \textit{a priori} the accuracy gained by a neural network is not satisfactory. 
%
%
%
%
\section{ChemTab: Jointly Learning the Progress Variables and Manifold Approximation}
As mentioned earlier, to reduce the computational effort in coupled simulations, state-space parametrization approaches follow a two-phase strategy, first, parametrize and tabulate {\em a priori} the scalar evolution of a reactive turbulent environment by few progress variables that govern the scalar evolution in a laminar flame, and second, use a tabular lookup at run-time to determine the high-dimensional chemical state required by the CFD solver. For instance, the FGM approach replaces all species and temperature by a {\em mixture-fraction} and a single {\em reaction progress variable} or reaction progress parameter. In this study, we focus on state-space parametrization using {\em Unsteady Flamelet Generated Manifolds} or Unsteady FGMs~\cite{bojko2016}. We modify this approach in three ways: the progress variable generation is different, the manifold is not tabulated and lastly, the progress variables and manifold approximation are done jointly.

\subsection{Background: Unsteady FGM}
FGM is a widely used tabulated chemistry method and can deal with a range of complicated conditions. FGM model shares the same theoretical basis with flamelet approaches~\cite{PETERS1984319}, in which a multi-dimensional flame can be considered as an ensemble of multiple one-dimensional flames. Generally FGM model used for combustion modeling follows three steps as shown below:
\begin{enumerate}
    \item Calculation of the representative 1-D flamelets.
    \item Transformation of 1-D flamelets solutions to progress variables space.
    \item Retrieval of thermo-chemical variables from the FGM tables according to FGM control variables from CFD simulations.
\end{enumerate}
\subsubsection{Governing Equations}\label{subsec:1Dflamesol}
\begin{table}
  \caption{Definitions for terms used in Section~\ref{subsec:1Dflamesol}}
  \label{tab:maesource}
  \begin{tabular}{ll}
   \makecell {
        \begin{tabular}{|l|l|}
            \hline
            Symbol & Description\\
            \hline
            $Z_{mix}$ & Mixture Fraction\\
            $C_{pv}$ & Progress Variable\\
            $Y$ & Species Mass Fraction\\    
            $\dot{S}$ & Species Source Terms\\
            $\rho$ & Density of the mixture\\
            $T$ & Temperature of the mixture\\
            $\dot{S_{i}}$ & \makecell {Source Term of the \\$i^{th}$ species}\\
            $h^0_{f,i}$ & \makecell {Heat of Formation of the \\$i^{th}$ species}\\
            $\mathcal{D}_{i}$ & Diffusivity of $i^{th}$ species\\
            $\kappa$ & Thermal Conductivity\\
         	$Pr$ & Prandtl number\\
            $Sc$ & Schmidt number\\
            \hline
        \end{tabular}
    }\quad\quad\quad &
    \quad\quad\quad
    \makecell {
        \begin{tabular}{|l|l|l|l|l|}
            \hline
             Symbol & Description\\
            \hline
            $Le$ & Lewis number \\
            $\mu$ & Viscosities \\
            $h$ & Total Enthalpy\\
            $s$ & Total no. of Species in Mechanism\\
            $p$ & Number of Progress Variables\\
            $n$ & Number of Data Points\\
            $\widehat{Y}$ & Reactive Scalars\\
            $ \widehat{\dot{S}}$ & Reactive Scalars Source Terms\\
            $\phi$ & Non-Linear function of $Y$\\
            $\zeta$ & Non-Linear function of $\widehat{Y}$\\
            $\psi$ & Non-Linear function of $\widehat{Y}$ and $Z_{mix}$\\
            $k$ & \makecell{No. of Species used to \\generate Progress Variables}\\
           \hline
        \end{tabular}
    }\quad\quad   
  \end{tabular}
\end{table}
Conservation equations for mass, species, momentum and energy for the 1-D, fully compressible, and viscous flames, are given by:
\begin{eqnarray}
\frac{\partial \rho}{\partial t} + \frac{\partial \left(\rho u_x \right)}{\partial x} & = & 0\\ \label{eqn:1Dflamea}
    \frac{\partial \left( \rho Y_i \right)}{\partial t} + \frac{\partial \rho u_x Y_i}{\partial x} & = & \frac{\partial}{\partial x}\left( \rho \mathcal{D}_i \frac{\partial Y_i}{\partial x} \right) + \dot{S_i} \\\label{eqn:1Dflameb}
    \frac{\partial \left (\rho u_x \right)}{\partial t} + \frac{\partial \left( \rho u_x^2 \right)}{\partial x} & = & -\frac{\partial p}{\partial x} + \frac{\partial}{\partial x} \left(\mu \frac{\partial u_x}{\partial x} \right) \\\label{eqn:1Dflamec}
    \frac{\partial \left(\rho e_t \right)}{\partial t} + \frac{\partial}{\partial x} \left(\rho u_x H_t \right) & = & \frac{\partial}{\partial x} \left(u_x \mu \frac{\partial u_x}{\partial x} \right) + \mu \frac{c_p}{Pr} \left( 1 - \frac{1}{Le} \right)\frac{dT}{dx}\\ \label{eqn:1Dflamed}
            & + &\frac{1}{Sc} \frac{dh}{dx} - \sum \dot{S_i} h^o_{f,i}\nonumber
\end{eqnarray}
where the different terms are defined in Table~\ref{tab:maesource}.


We simplify the above equations making some well known assumptions. In 1D cartesian coordinates, the steady state solution to~\eqref{eqn:1Dflamea}--\eqref{eqn:1Dflamed} is obtained only when the total mass flux is zero, i.e., velocity field is zero $(u_x = 0)$ and so the four equations reduce to:
\begin{subequations}\label{eqn:energysteadymain}
    \begin{alignat}{3}
        &\frac{\partial}{\partial x} \left( \rho \mathcal{D}_{i} \frac{\partial Y_i}{\partial x} \right) + \dot{S_i} = 0 \label{eqn:speciessteady} \\
        &\frac{\partial}{\partial x} \left(\kappa \frac{\partial T}{\partial x} + \sum \rho \mathcal{D}_{i} \frac{\partial Y_i}{\partial x} h_i \right) - \sum \dot{S_i} h^o_{f,i} = 0. \label{eqn:energysteady}
    \end{alignat}
\end{subequations}

In~\eqref{eqn:energysteady}, the final term in the energy equation is represented by the total sum of the product of all the source species and their respective heat of formation and is collectively called the source energy. Source energy is one of the crucial parameters in the combustion simulation and accurate chemistry description is required to define it. Prediction error of this term is used as the basis of comparison of our method against the other state of the art methods.

\subsubsection{Flamelet Solutions}
The data is generated by solving 1-D Steady State Flamelets differential equations in~\ref{eqn:energysteady} using a finite volume PDE solver. The species $Y$ and thermochemical state variables $\dot{S}$ are generated using the solver.

\begin{subequations}
\label{eqn:datagenerated}
    \begin{alignat}{3}
        Y = \begin{bmatrix}
            Y_{11} &..&..&Y_{1s}\\
            ..&..&..&..\\
            ..&..&..&..\\
            ..&..&..&..\\
            Y_{n1}&..&.. & Y_{ns}
        \end{bmatrix},\quad
        \dot{S} = \begin{bmatrix}
            S_{11} &..&..&S_{1s}\\
            ..&..&..&..\\
            ..&..&..&..\\
            ..&..&..&..\\
            S_{n1}&..&.. & S_{ns}
        \end{bmatrix},\quad
        Z_{mix} = \begin{bmatrix}
            Z_{mix_{1}}\\
            ..\\
            ..\\
            ..\\
            Z_{mix_{n}}
        \end{bmatrix}
    \end{alignat}
\end{subequations}

\subsection{ChemTab}
In ChemTab, the unsteady FGM approach is replaced with the following three steps:
\begin{enumerate}
    \item Calculation of the representative 1D flamelets (data generation)
    \item Using the data generated jointly generate Progress Variables  (encoder) and Manifold Approximation (regressor) using ChemTab
    \item Retrieval of thermo-chemical variables from the ChemTab-regressor according to progress variables from CFD simulations.
\end{enumerate}

\subsubsection{Formulation}
The generated data described in ~\eqref{eqn:datagenerated} is then used by ChemTab. Conceptually the following equations summarize the relationships:

\begin{subequations}\label{eqn:conceptual-energysteady}
    \begin{alignat}{3}
        \dot{S} = \phi(Y)\\
        S_{energy} = - \sum_{i}^{s} h_{f,i}^{0} * \dot{S_{i}}
    \end{alignat}
\end{subequations}

The two sub-problems of state-space parametrization are formulated as a joint optimization problems as follows: 
\begin{subequations}\label{eqn:chemtabformulation}
    \begin{alignat}{3}
        min (\sum_{i=1}^{k}\sum_{j=1}^{n} || {\dot{S_{ij}}} - \zeta_{i}(\widehat{Y_{j}}) ||_{t} + \sum_{j=1}^{n} ||S_{energy} - \psi(\widehat{Y}, Z_{mix})||_{t})\\
        \textrm{s.t.} \quad t \in R \\
            \def\sss{\scriptscriptstyle}
            \setstackgap{L}{8pt}
            \def\stacktype{L}
            \stackunder{\mathrm{\widehat{Y}}}{\sss n\times p} =  \stackunder{Y}{\sss n\times s} \times \stackunder{W}{\sss s\times p} \\
            p << s\\
            ||W||=1 \label{eqn:UN-constraint} \\
            W^{T} \times W = I \label{eqn:WO-constraint} \\
            (\widehat{Y} \oplus Z{mix})^{T} \times (\widehat{Y} \oplus Z{mix})= I \label{eqn:AR-constraint} \\
        \dot{S} \approx \widehat{\dot{S}} = \zeta(\widehat{Y}, Z_{mix})\\
        S_{energy} \approx \widehat{S_{energy}} = \psi(\widehat{Y}, Z_{mix})
   \end{alignat}
\end{subequations}
The formulation described in equation \ref{eqn:chemtabformulation} learns the optimal reactive scalars $C_{pv}s$ (described by the embedding $Y \times W$) that along with $Z_{mix}$ form the progress variables. This is a linear dimensionality reduction problem such that the new basis retains the inherent physics in higher dimensions described by the non-linear relation between $Y$ and $\dot{S}$. To facilitate the development of transport equations using the progress variables it is necessary that the embedding of the variables in the low-dimensional space be linear. The constraints on the linear embedding are inspired by the work of \cite{ihme20127715} and the key ideas from PCA. 


\subsubsection{Implementation}\label{sec:ChemTabImplementation}

The joint optimization problem is solved using a Deep Neural Architecture.
ChemTab jointly optimizes two neural networks for the tasks of reaction progress variable generation ({\em encoder}) and manifold approximation ({\em regressor}). The encoder network focuses on linear dimensionality reduction and creates a linear embedding for the input. The regressor network focuses on learning the manifold approximation: a regression function whose input is the linear embedding and the output are the desired thermo-chemical state variables.

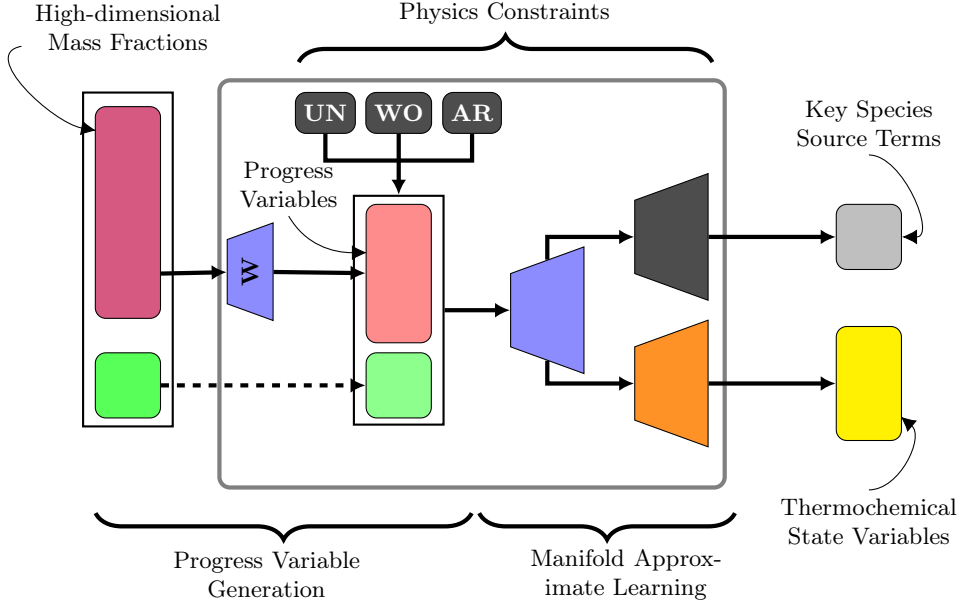
\begin{figure*}[h]
  \centering
  \begin{tikzpicture}
\tikzset{
  >={Latex[width=2mm,length=2mm]},
  base/.style={draw, on grid, align=center, minimum height=4ex},
  proc/.style={base, rectangle, text width=2em, text height=5em},
  term/.style={proc, rounded corners, text width=2em, text height=8em},
  cons/.style={base, rounded corners, text=white, fill=black!70}
}
\node[term,fill=purple!65] (input) {};
\node [below,yshift=4em,text width=8em,align=center,yshift=0.5em] at (input.north) (inputtxt){High-dimensional Mass Fractions};
\node[term,text height=2em,below of=input,yshift=-4em,fill=green!65] (zmix) {};
\node[base,thick,text height=13em,above of=input,text width=3em,yshift=-5em] (inputi){};
\node[term,fill=red!45,text height=5em,right of=input,xshift=8em,yshift=-2.5em] (linear) {};
\node[term,text height=2em,right of=zmix,xshift=8em,fill=green!45] (zmixi) {};
\node[base,thick,text height=8.7em,right of=input,xshift=8em,text width=3em,yshift=-4em] (lineari){};
\node[left of=lineari,xshift=-6.5em] (inputii) {};
\node (nnlinear) [draw,trapezium,trapezium left angle=70,trapezium right angle=70,minimum width=4em, minimum height=0.5em,rotate=90,right of=inputii,xshift=-1.5em,yshift=-3.5em,fill=blue!45] {${\bf W}$};
\node (nn) [draw,trapezium,trapezium left angle=70,trapezium right angle=70,minimum width=5em, minimum height=3em,rotate=90,below of=lineari,yshift=-3em,fill=blue!45] {};
\node (nn1) [draw,trapezium,trapezium left angle=70,trapezium right angle=70,minimum width=5em, minimum height=3em,rotate=90,below of=nn,yshift=-2em,xshift=3em,fill=black!70] {};
\node (nn2) [draw,trapezium,trapezium left angle=70,trapezium right angle=70,minimum width=5em, minimum height=3em,rotate=90,below of=nn,yshift=-2em,xshift=-3em,fill=orange!85] {};
\node[base,ultra thick,rounded corners,draw=gray,text height=16em,right of=inputi,xshift=11em,text width=20em,yshift=-1em] (nni){};
\node[term,fill=yellow,right of=nn2,xshift=5em,text height=4em] (output) {};
\node [below,text width=8em,yshift=-2em,align=center] at (output.south) (outputtxt){Thermochemical State Variables};
\node[term,fill=black!25,right of=nn1,xshift=5em,text height=2em] (source) {};
\node [above,text width=8em,yshift=2em,align=center] at (source.north) (sourcetxt){Key Species Source Terms};
\coordinate[xshift=0em,yshift=-2.5em] (inputc) at (input.east);
\draw[->,ultra thick]  (inputc) -- (nnlinear);
\draw[->,ultra thick]  (nnlinear) -- (linear);
\draw[->,ultra thick,dashed]             (zmix) -- (zmixi);
\draw[->,ultra thick]             (lineari) -- (nn);
\draw[->,ultra thick]             (nn2) -- (output);
\draw[->,ultra thick]             (nn1) -- (source);
\draw[->,ultra thick] (nn) |- (nn1);
\draw[->,ultra thick] (nn) |- (nn2);
\coordinate[xshift=-1.3em,yshift=-1em] (st) at (input.north);
\draw[->] (inputtxt) to [out=190,in=170] (st);
\coordinate[xshift=1.3em,yshift=1em] (st1) at (output.south);
\draw[->] (outputtxt.north) to [out=0,in=-40] (st1);
\draw[->] (sourcetxt.south) to [out=0,in=-0] (source.east);
\node [above,text width=4.5em,xshift=-4.5em,yshift=-0.5em,align=center] at (linear.north) (pvtxt){Progress Variables};
\coordinate[xshift=-1.3em,yshift=-2em] (st2) at (linear.north);
\draw[->] (pvtxt.south) to [out=-60,in=180] (st2);
\node[cons, above of=lineari,yshift=5em](wo){\bf WO};
\node[cons, above of=lineari,xshift=-3em,yshift=5em](un){\bf UN};
\node[cons, above of=lineari,xshift=3em,yshift=5em](ar){\bf AR};
\coordinate[yshift=1.4em] (st3) at (lineari.north);
\draw[->,ultra thick] (wo) to (lineari.north);
\draw[-,ultra thick] (un) |- (st3);
\draw[-,ultra thick] (ar) |- (st3);
\draw [decorate,decoration={brace,amplitude=10pt},ultra thick,xshift=2pt]
  (2.2,2) -- (7.7,2) node [xshift=-8.5em,yshift=2em] 
  {Physics Constraints};
\draw [decorate,decoration={brace,amplitude=10pt},ultra thick,xshift=2pt]
  (4.5,-4) -- (-0.5,-4) node [text width=8em,xshift=7em,yshift=-2.5em,align=center] 
  {Progress Variable Generation};
\draw [decorate,decoration={brace,amplitude=10pt},ultra thick,xshift=2pt]
  (8,-4) -- (4.6,-4) node [text width=8em,xshift=6em,yshift=-2.5em,align=center] 
  {Manifold Approximate Learning};
\end{tikzpicture}
  \caption{ChemTab Architecture}
\end{figure*}

\begin{table}
  \caption{Symbols used in Section \ref{sec:ChemTabImplementation}}
  \label{tab:chemtabdnnsymbols}
  \begin{tabular}{ll}
   \makecell {
        \begin{tabular}{|l|l|}
            \hline
            Symbol & Description\\
            \hline
            $f_{\theta}$ & Prediction Function \\
            $y$ & Input/Output Matrix \\
            $W$ & Weight Matrix \\
            $b$ & Bias Matrix \\
            $\mathcal{S}$ & Themochemical state variables\\ 
            $\mathcal{\sigma}$ &scalar/activation function \\
            $\rm o$ & Entry-wise operation\\
            \hline
        \end{tabular}
    }\quad\quad\quad &
    \quad\quad\quad
    \makecell {
        \begin{tabular}{|l|l|l|l|l|}
            \hline
             Symbol & Description\\
            \hline
         	$m$ & Number of neurons\\
            $n$ & Number of Data Points\\
            $in$ & $s \times n$ \\
            $s$ & Total no. of Species in Mechanism\\
            $out$ & No. of thermo-chemical variables\\
            $d_{in}$ & Input Dimensions $s$\\
            $d_{out}$ & Output Dimensions $s+1$ \\
            $L$ & No. of Layers\\
           \hline
        \end{tabular}
    }\quad\quad   
  \end{tabular}
\end{table}

\begin{equation}\label{eqn:DNN}
    \begin{aligned}
    \begin{split}
      f_{\mathcal{\theta}}(y) = W^{[L-1]}\mathcal{\sigma} \: \mathcal{\rm o} \: (W^{[L−2]}\mathcal{\sigma} \: \mathcal{\rm o} \: (\dots(W^{[1]}\mathcal{\sigma} \: \mathcal{\rm o} \\
           \: (W^{[0]}y + b^{[0]}) + b^{[1]})\dots) + b
    \end{split}\\
     \textrm{where,} \quad W^{[l]} \in R^{m_{l+1} \: \times \: m_{l}}\\
      b^{[l]} = R^{m_{l+1}}\\
      m_{0} = d_{in} = d\\
      m_{L} = d_{out}
    \end{aligned}
\end{equation}
As described by~\eqref{eqn:DNN} a Deep Neural Network can be conceptualized as a series of operations. The input of the network is the data for each of the species for each flame at each axial coordinate. 

\begin{equation}\label{eqn:DNN-expansion}
    \begin{aligned}
    f_{\mathcal{\theta}}^{[0]}(y) = y \\
    f_{\mathcal{\theta}}^{[1]}(y) = (W^{[0]} f_{\mathcal{\theta}}^{[0]}(y)) \\
        f_{\mathcal{\theta}}^{[2]}(y) = (f_{\mathcal{\theta}}^{[2]}(y) \oplus Z_{mix}) \\
    f_{\mathcal{\theta}}^{[l]}(y) = \mathcal{\sigma} \: \mathcal{\rm o} \: (W^{[l−1]} f_{\mathcal{\theta}}^{[l-1]}(y) \:+\: b^{[l-1]} ) \:\:\: \forall \quad l \quad \textrm{s.t.} \quad {3 \geq l \leq L-1}  \\
    f_{\mathcal{\theta}}(y) = f_{\mathcal{\theta}}^{[L]}(y) = \mathcal{\sigma} \: \mathcal{\rm o} \: (W^{[L-1]} f_{\mathcal{\theta}}^{[L-1]}(y) \:+\: b^{[L-1]} ) 
    \end{aligned}
\end{equation}
As described by~\eqref{eqn:DNN-expansion} the network is a layer-wise composition. The input of the network is reduced at the first layer linearly: this creates the linear embedding/reacting scalars ($C_{pv}s$). The next layer concatenates the conserved scalar $Z_{mix}$ with the reacting scalars. These progress variables are then fed to the next layer. The subsequent layers together make up the regressor that learns a non-linear function between the progress variables and the thermo-chemical state variables.

\begin{equation}\label{eqn:ChemTab-Constraints}
    \begin{aligned}
        \arg\min_{\mathcal{\theta}} \quad \lvert f_{\mathcal{\theta}}(y) - \mathcal{S} \rvert\\
        s.t. \quad W^{[0]T}W^{[0]} = I\\
        \Vert W^{[0]} \Vert = 1\\
        f_{\mathcal{\theta}}^{[2]}(y)^Tf_{\mathcal{\theta}}^{[2]}(y) = I
    \end{aligned}
\end{equation}
 
As described by~\eqref{eqn:ChemTab-Constraints}, ChemTab minimizes the Mean Absolute Error in predicting the thermo-chemical state variables (Source Energy in the current work) while ensuring that the linear embedding conforms to the following constraints:
\begin{enumerate}\label{enu:encoder-conditions}
    \item Embedding Weights $w$ learnt are unit norm (UN)
    \item Embedding Weights $w$ learned for the species mass fractions $Y_i$s are uncorrelated/orthogonal (WO)
    \item The reaction progress variables are uncorrelated/orthogonal (AR)
\end{enumerate}

The constraints in~\eqref{eqn:ChemTab-Constraints} will be also added to the objective in addition to the predictions of key source terms, corresponding to a few important species, which serve as the physics constraints. 
\subsubsection{Extensions}
The current framework and implementation can be very easily extended to include the prediction of additional thermochemical state variables and the projection of the embedding to get back the high dimensional mass fractions. These can be implemented as two other neural networks and their respective prediction errors can be added to the objective function. 
\section{Experimentation and Results}
In this section we explain the specifics of the data set used, the training strategy, impact of the number of $C_{pv}$, comparison with the existing framework and relevant machine learning methods and the performance of the best model in the context of the multiple objectives.

\subsection{Dataset}\label{sec:datageneration}
The training data was generated by solving 1-D Steady State Flamelets differential equations using a finite volume PDE solver. To model the chemical kinetics reaction rates, a variety of mechanisms are adopted in the combustion community. Depending on the hydrocarbon fuel different mechanisms are chosen which closely describe the chemistry associated with the fuel of simulation. Methane is the basic hydrocarbon and one of the major products of many higher order hydrocarbons. GRI-Mech 3.0 is one of the widely used Methane mechanism to model the reaction kinetics. This mechanism consists of 53 chemical species and 325 reactions. 

The Flamelet solver discretizes the domain into $200$ grid points (200 observations on the axial coordinate) in between the fuel and the air boundary and $100$ flame are solved to steady-state. Once the solution reaches steady-state the solver completes one iteration. For the next iteration flame solution is strained by reducing the domain by $0.99$ and the process is continued until the flame extinguishes. Each flame is then tagged with the corresponding strain rate that is called a flame-key. To train the model 20,000 data points (100 flames and 200 grid points) for a single pressure setting are used. The data is generated using an in-house solver which creates the flame solutions and stores the required data. Some of the generated data that represent extinguished flames were discarded, which led to exclusion of approximately 3,500 data points.

We experiment the model training and evaluation using two strategies:
\begin{inparaenum}[a.] 
\item {\em 50\% Flamelets} - Train using data from 50\% of flamelets selected randomly and test using data from the remaining 50\% of the flamelets, and, 
 \item {\em 50\% Data points} - Train using 50\% data points selected randomly, and test on remaining 50\% data.
\end{inparaenum}
\subsection{Evaluation}

We use the Mean Absolute Error of the Source Energy across the entire dataset as the metric to compare the performance of ChemTab against the all the other methods.

\begin{equation}
  MAE = \frac{1}{n}\sum_{i=0}^{n} \vert S - \widehat{S}\vert
\end{equation}

\subsection{Implementation and Settings}
We implemented ChemTab using Tensorflow 2.3.0, Keras and Adam optimizer. Models were trained on a server with Nvidia Quadro RTX 5000 GPU and cuDNN 8.0 and CUDA 11.0. We performed a coarse grid search on the hyperparameters (dropouts,learning rate, early stopping, batch size) \& standard model architecture (number of layers, number of nodes in the layers, activation functions). 
 \begin{table}\label{tbl:results}
 \centering
 {\small
  \caption{Model Parameters}
  \label{tab:ChemTabParameters}
  \begin{tabular}{|ll||ll|}
    \hline
    Parameter & Value & Parameter & Value\\
    \hline
 	Learning Rate & 0.001& Number of Layers  & 11\\
 	Momentum & 0.5 & Layer Shapes & 53,1|4|5|32|64|128|256|512|256|128|64|32|8 \\
    Dropout  & 5\% & Activation Functions & ReLU\\
    Early Stopping  & Yes &Number of epochs & 500 (short run) | 20000 (long run)\\
    Batch Size  & 32 & Network Weight Initialization & Uniform Distribution\\ 
   \hline
\end{tabular}}
\end{table}

After the initial model architecture and hyper-parameter search, all subsequent models in the subsequent studies were trained for 500 epochs. Finally, the reported results are taken as average of 10 runs.

\subsection{ChemTab Variants}
The constraints on the linear embedding are important architectural components and we measure the performance of the following variations of ChemTab against the framework and current state-of-the-art.

\begin{table}
  \caption{ChemTab Architectural Variants}
  \label{tab:archcomponents}
  \begin{tabular}{|l|c|}
    \hline
    Abbreviation & Description\\
    \hline
     UN\label{chemtab:un} & \makecell[l] {Unit Norm Constraint on Weights of the Linear Embedding} \\
     WO\label{chemtab:wo} & \makecell[l] {Orthogonality Constraint on Weights of the Linear Embedding}\\
     AR\label{chemtab:ar} & \makecell[l] {Orthogonality Constraint on Linear Embedding concatenated with $Z_{mix}$} \\
     UN + WO\label{chemtab:unwo} & \makecell[l] {Unit Norm Constraint and  
      Orthogonality Constraint on Weights of the Linear Embedding} \\ 
     UN + AR\label{chemtab:unar} & \makecell[l] {Unit Norm Constraint on the 
      Weights and \\Orthogonality Constraint 
      on Linear Embedding concatenated with $Z_{mix}$ }\\
     WO + AR\label{chemtab:woar} & \makecell[l] {Orthogonality Constraint on the
      Weights and Linear Embedding concatenated with $Z_{mix}$} \\ 
     All\label{chemtab:all} & \makecell[l] {Unit Norm and Orthogonality Constraint on the\\
      Weights and Linear Embedding concatenated with $Z_{mix}$} \\
  \hline
\end{tabular}
\end{table}

\subsection{Compared Methods}
We compare the 7 variants of ChemTab with the relevant constraints on the Linear Embedding and the Progress Variables with a series of state-of-the-art baselines for Source Energy prediction. In the table \href{tab:baselines} find the related machine learning methods discussed in the sections \href{sec:relatedwork}.

\begin{table}
\centering
  \caption{Current state of the art methods and ChemTab variants}
  \label{tab:baselines}
  \begin{tabular}{|l|c|l|}
    \hline
    Method Abbreviation & Progress Variable Generation & Manifold Approximation\\
   \hline
    FGM-CPVG-DNN\label{method:FGM-CPVG-DNN} & \makecell[l] {FGM Constrained  }  & \makecell {DNN}\\
    PCA-PVG-DNN\label{method:PCA-PVG-DNN} & \makecell[l] {PCA }  & \makecell {DNN}\\
    DNN-PVG(NL)-DNN\label{method:DNN-PVG(NL)-GP} & \makecell[l] {Non-Linear Encoder}  & \makecell {DNN}\\
    DNN-PVG(UL)-DNN\label{method:DNN-PVG(UL)-DNN} & \makecell[l] {Unconstrained Linear Encoder }  & \makecell {DNN}\\
    CT-PVG(ALL)-DNN\label{method:CT-PVG(ALL)-DNN} & \makecell[l] {Physics constrained Linear Encoder \ref{chemtab:all}}  & \makecell {DNN}\\
    CT-PVG(UN)-DNN\label{method:CT-PVG(UN)-DNN} & \makecell[l] {Physics constrained (UN)  Linear Encoder \ref{chemtab:un}}  & \makecell {DNN}\\
    CT-PVG(WO)-DNN\label{method:CT-PVG(WO)-DNN} & \makecell[l] {Physics constrained (WO) Linear Encoder \ref{chemtab:wo}}  & \makecell {DNN}\\
    CT-PVG(AR)-DNN\label{method:CT-PVG(AR)-DNN} & \makecell[l] {Physics constrained (AR) Linear Encoder \ref{chemtab:ar}}  & \makecell {DNN}\\
    CT-PVG(UN+WO)-DNN\label{method:CT-PVG(UN+WO)-DNN} & \makecell[l] {Physics constrained (UN+WO)  Linear Encoder \ref{chemtab:unwo}}  & \makecell {DNN}\\
    CT-PVG(UN+AR)-DNN\label{method:CT-PVG(UN+AR)-DNN} & \makecell[l] {Physics constrained (UN+AR) Linear Encoder \ref{chemtab:unar}}  & \makecell {DNN}\\
    CT-PVG(WO+AR)-DNN\label{method:CT-PVG(WO+AR)-DNN} & \makecell[l] {Physics constrained (WO+AR) Linear Encoder \ref{chemtab:woar}}  & \makecell {DNN}\\
\hline
\end{tabular}
\end{table}

\subsection{Results}

\subsubsection{Current Framework Comparison}
The current framework uses FGM based progress variables and Conformal Mapping based Tabulation and Lagrange Polynomial Interpolation based lookup. The tabulation was generated by using the entire data-set. The best MAE that the framework generated on the data-set was 2.243 E+09. The best ChemTab model trained on 50\% of the data showed a 73\% reduction in error. This reduction although high comes from the limitation of the current framework to include more than 2 progress variables and the realization of that through conformal mapping. We present a more principled comparison with the state-of-the-art methods in the next section.

\subsubsection{Other Baseline Comparisons}
We include {\em DNN-PVG(NL)-DNN} although cannot be used due to non-linear embedding serves as a benchmark. We also compared ChemTab to Gaussian Processes and we performed better 10\%, however there are several challenges with operationalization of Gaussian Process in our context and so we focus more on bench-marking against the relevant DNN based approaches.

\begin{figure}[h]
  \centering
  \caption{MAE for Source Energy: Data Set Split Strategy}
  \begin{tikzpicture}
\begin{axis}[
xbar, xmin=0,
xlabel={MAE},
symbolic y coords={{FGM-CPVG-DNN},{PCA-PVG-DNN},{DNN-PVG(NL)-DNN},{DNN-PVG(UL)-DNN},{CT-PVG(UN)-DNN},{CT-PVG(WO)-DNN},{CT-PVG(AR)-DNN},{CT-PVG(UN+WO)-DNN},{CT-PVG(UN+AR)-DNN},{CT-PVG(WO+AR)-DNN},{CT-PVG(ALL)-DNN}},
ytick=data,
nodes near coords, nodes near coords align={horizontal},
ytick=data,
]
\addplot 
	coordinates {
	(2.93E+09,{FGM-CPVG-DNN}) (8.99E+08,{PCA-PVG-DNN}) (8.90E+08,{DNN-PVG(NL)-DNN}) (7.62E+08,{DNN-PVG(UL)-DNN}) (7.78E+08,{CT-PVG(UN)-DNN}) (7.66E+08,{CT-PVG(WO)-DNN}) (8.12E+08,{CT-PVG(AR)-DNN}) (7.05E+08,{CT-PVG(UN+WO)-DNN}) (8.05E+08,{CT-PVG(UN+AR)-DNN})  (6.01E+08,{CT-PVG(WO+AR)-DNN}) (5.58E+08,{CT-PVG(ALL)-DNN})    
	};
\addplot 
	coordinates {(1.13E+10,{FGM-CPVG-DNN}) (5.28E+09,{PCA-PVG-DNN}) (9.01E+09,{DNN-PVG(NL)-DNN}) (5.47E+09,{DNN-PVG(UL)-DNN}) (3.94E+09,{CT-PVG(UN)-DNN}) (4.89E+09,{CT-PVG(WO)-DNN}) (9.26E+09,{CT-PVG(AR)-DNN}) (5.60E+09,{CT-PVG(UN+WO)-DNN}) (5.62E+09,{CT-PVG(UN+AR)-DNN}) (3.96E+09,{CT-PVG(WO+AR)-DNN}) (3.04E+09,{CT-PVG(ALL)-DNN})

};
\legend{50\% Data,50\% Flamelets}
\end{axis}
\end{tikzpicture}
  \label{fig:datasetsplitresults}
\end{figure}

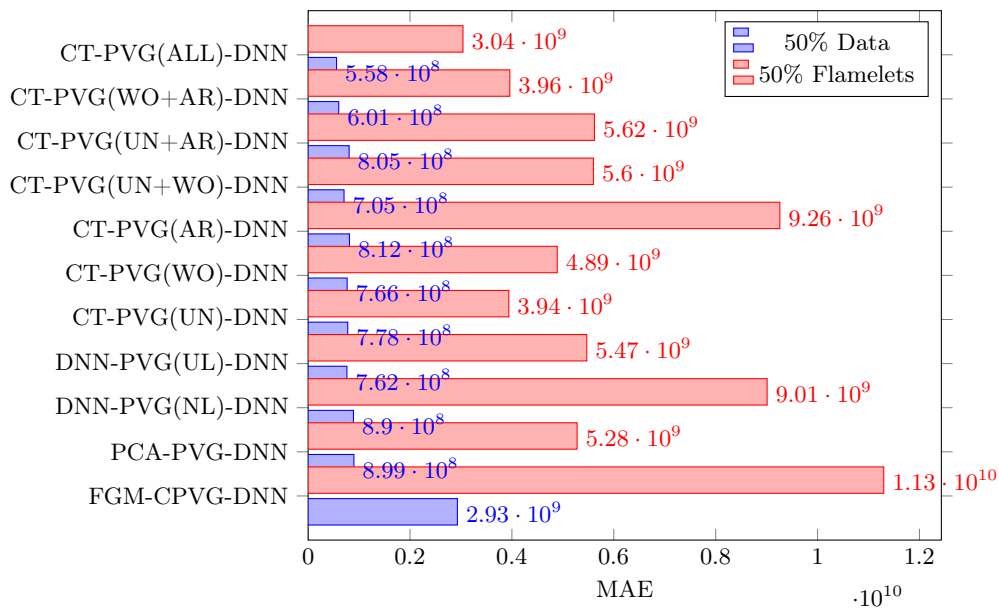
Figure \ref{fig:datasetsplitresults} shows the results of an ablation study for both types of sampling strategies. When the trained using the sampled points, all models consistently do better than when trained using sampled flamelets. Essentially the flame is considered as an ensemble of multiple one-dimensional flamelets, each of which captures some of the highly nonlinear state-space and hence almost all models struggle in this training regime. ChemTab models still perform better and our assertions are that our constraints help in the generalization process. Our dataset is limited and so we limit ourselves to use only 50\% of the data for training.

\begin{figure}[h]
  \centering
  \caption{MAE for Source Energy: Cpv Ablation}
  \begin{tikzpicture}
\begin{axis}[
    title={MAE by \# $C_{pv}$},
    xlabel={\#$C_{pv}$},
    ylabel={MAE},
    xmin=1, xmax=5,
    ymin=5.40E+08, ymax=2.4E+09,
    xtick={1,2,3,4,5},
    ytick={5.40E+08,7.40E+08,9.40E+08,2.40E+09,4.40E+09,6.40E+09, 8.40E+09},
    legend pos=outer north east,
    ymajorgrids=true,
    grid style=dashed,
]

\addplot[color=blue!50,mark=square,]
    coordinates {
    (1,5.33E+09)	(2,1.56E+09)	(3,1.01E+09)	(4,8.99E+08)	(5,1.26E+09)
    };
    \addlegendentry{PCA-PVG-DNN}
\addplot[color=red!50,mark=triangle,]
    coordinates {
    (1,1.16E+09)	(2,8.32E+08)	(3,7.48E+08)	(4,8.90E+08)	(5,7.25E+08)
    };
    \addlegendentry{DNN-PVG(NL)-DNN} 
\addplot[color=cyan,mark=pentagon,]
    coordinates {
    (1,1.06E+09)	(2,8.35E+08)	(3,1.12E+09)	(4,7.62E+08)	(5,6.21E+08)
    };
    \addlegendentry{DNN-PVG(UL)-DNN}     
    
\addplot[color=magenta,mark=diamond*,]
    coordinates {
    (1,1.00E+09)	(2,7.72E+08)	(3,7.25E+08)	(4,5.58E+08)	(5,6.00E+08)
    };
    \addlegendentry{CT-PVG(ALL)-DNN}  
    
\end{axis}
\end{tikzpicture}
  \label{fig:cpvablationresults}
\end{figure}
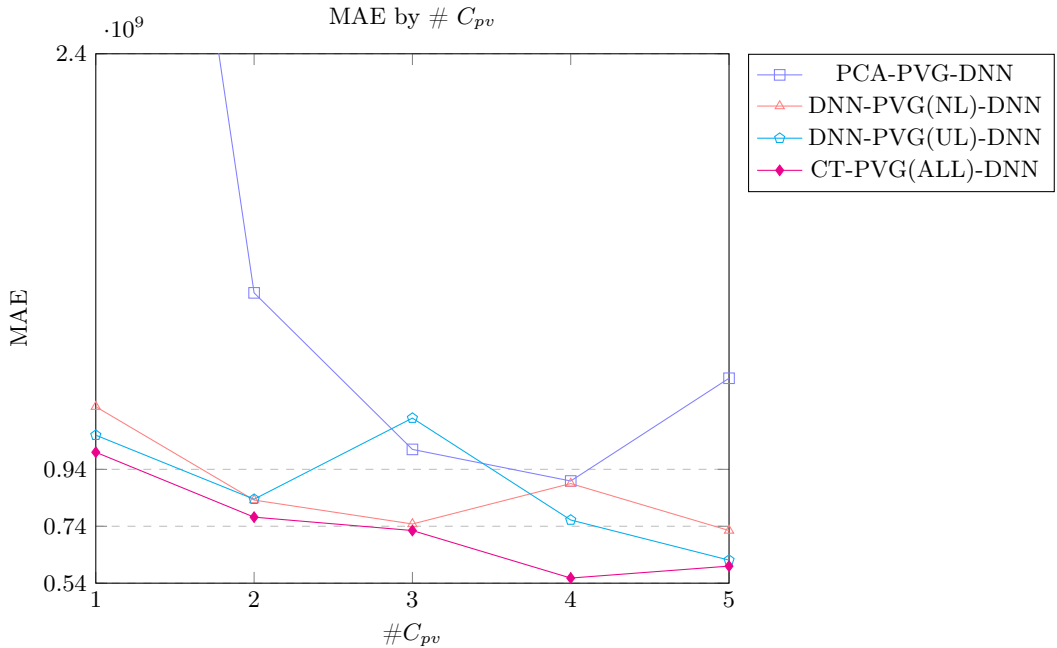
As we increase the number of $C_{pv}$ the computational time of the flow simulation goes up, so we want to use the least number of $C_{pv}$ while still capturing the essential physics. Figure \ref{fig:cpvablationresults} shows the MAE decreases with increase in the number of $C_{pv}$ and then starts to increase again. As we add more $C_{pv}$ the embedding has too many degrees of freedom and hence may start diverging.

\subsubsection{Best Model Performance}

\begin{figure}[h]
  \centering
  \caption{MAE for Key Source Terms - Best Model}
  \begin{tikzpicture}[scale=0.7]
\begin{axis}[ 
xbar, xmin=0,
xlabel={MAE},
symbolic y coords={{Source Species - O2},{Source Species - CO},{Source Species - CO2},{Source Species - H2O},{Source Species - OH},{Source Species - H2},{Source Species - CH4}},
ytick=data,
nodes near coords, 
nodes near coords align={horizontal},
ytick=data,
]
\addplot coordinates {
    (2.1896E+02,{Source Species - O2})
    (3.1807E+01,{Source Species - CO})
    (1.9002E+01,{Source Species - CO2})
    (3.5005E+01,{Source Species - H2O})
    (5.5288E+01,{Source Species - OH})
    (6.0132E+00,{Source Species - H2})
    (2.3982E+01,{Source Species - CH4})
    };
\end{axis}
\end{tikzpicture} 
  \label{fig:keysourcetermsresults}
\end{figure}

 \begin{table}
  \caption{Constraints - Best Model}
  \label{tab:constraints}
  \begin{tabular}{lll}
   \\
   \makecell{Unit Norm Constraint\\
                on the Weights of the\\
       Linear Embedding}
    & 
    \makecell{Orthogonality Constraint\\
                on the Weights of the\\
        Linear Embedding} 
    & 
    \makecell{Orthogonality Constraint\\
       on the Output of the\\
       Linear Embedding}
   \\\\
   \makecell {
        \begin{tabular}{|l|l|l|l|}
            \hline
             $w_{1}$ & $w_{2}$ & $w_{3}$ & $w_{4}$\\
            \hline
         	1.004 &1.005 &1.001 &0.998\\
           \hline
        \end{tabular}
    } &
    \quad
    \makecell {
        \begin{tabular}{|l|l|l|l|l|}
            \hline
              & $w_{1}$ & $w_{2}$ & $w_{3}$ & $w_{4}$\\
            \hline
         	$w_{1}$ &1.004 &-0.003 &-0.002 &0.005\\
            $w_{2}$ &-0.003 &1.004 &-0.003 &0.002\\
            $w_{3}$ &-0.002 &-0.003 &1.001 &0.001\\
            $w_{4}$ &-0.005 &0.002 &0.001 &0.99\\
           \hline
        \end{tabular}
    } & 
  \quad
    \makecell {
      \begin{tabular}{|l|l|l|l|l|l|}
        \hline
          & $Z_{mix}$& $Cpv_{1}$ & $Cpv_{2}$ & $Cpv_{3}$ & $Cpv_{4}$\\
        \hline
        $Z_{mix}$ &0.004 &0.00 &0.00 &0.00 &0.00\\
     	$Cpv_{1}$ &0.00 &0.008 &-0.001 &0.001 &0.00\\
        $Cpv_{2}$ &0.00 &-0.001 &0.008 &0.00 &0.00\\
        $Cpv_{3}$ &0.00 &0.001 &0.00 &0.007 &-0.001\\
        $Cpv_{4}$ &0.00 &0.00 &0.00 &-0.001 &0.067\\
       \hline
    \end{tabular}    
    }
        
  \end{tabular}
\end{table}

Table \ref{tab:constraints} shows the conformity of the constraints of equation \ref{eqn:chemtabformulation}. The first tabulation shows the \ref{eqn:UN-constraint} constraint conformity. The second tabulation shows the \ref{eqn:WO-constraint} constraint conformity and the third \ref{eqn:AR-constraint} constraint conformity. The \ref{eqn:AR-constraint} is also adequately satisfied as the constraint conformity is measured through covariance.

\subsubsection{Best Model Long Run Performance} 
We trained best model architecture on a 50\% Data Points strategy for a long run  of 20000 epochs and generated a MAE of 1.80E+08. Although counter-intuitive, this may be explained by double descent. Interested reader is referred to \cite{nakkiran2019deep} 
\section{Conclusion}
We propose ChemTab, a novel framework for jointly learning the progress variables and the manifold approximation for solving the high-dimensional chemistry in combustion models. In spirit, ChemTab follows the principle of physics guided neural networks~\cite{karpatne2017}, which are increasingly becoming popular for many scientific modeling tasks, though no solutions exist that can directly benefit the combustion community. ChemTab outperforms the state-of-the-art state-space parametrization in combustion. Crucially, ChemTab generated reaction progress variables can be interpreted by examining the weight matrix, {\em $W$}, and thus, allow for physical insights into the systems being modeled.

Incorporation of ChemTab into a flow simulation, as illustrated in Figure~\ref{fig:reduced}, will require prediction of additional thermochemical state variables and the projection of the embedding to get back the high dimensional mass fractions, and will be explored as part of future work.
\section{Acknowledgments}
Funded by the United States Department of Energy’s (DoE) National Nuclear Security Administration (NNSA) under the Predictive Science Academic Alliance Program III (PSAAP III) at the University at Buffalo, under contract number DE-NA0003961.
%
%
%
%
 \bibliographystyle{splncs04}

\bibliography{paper}

\end{document}